\documentclass[10pt]{article}
\usepackage[utf8]{inputenc}
\usepackage{times}
\usepackage{graphicx}
\usepackage{hyperref}
\usepackage{amsmath, amssymb}
\usepackage{caption}
\usepackage{subcaption}
\usepackage{tabularx}
\usepackage{booktabs}
\usepackage{algorithm}
\usepackage{algpseudocode}
\usepackage{url}
\usepackage{float}

\title{MAPS: Modeling Co-Existing Subjective Perspectives and Shared Meaning in Multi-Agent Cognitive Dialogue}

\date{
    \vspace{1em}
    \large \textit{Preprint – This is a preliminary version. A formal submission is in progress.}
}

\author{
  Molood Arman \\
  \texttt{arman.molood@gmail.com}
  \and
  Clément Bonnafous \\
  \texttt{cbonnafo31@gmail.com}
}

\begin{document}
\maketitle

\begin{abstract}
Human dialogue involves more than exchanging information—it expresses beliefs, emotions, and subjective cognitive styles. Yet current AI dialogue systems often enforce semantic uniformity, sacrificing diversity and interpretability. We present \textbf{MAPS (Multi-Agent Perspective Spaces)}, a novel framework that models dialogue between cognitively distinct agents through domain-weighted profiles, dynamic GRU-based memory, and interpretable token-level attention. MAPS enables agents to maintain individualized reasoning while progressively converging on shared meaning. Evaluations on \textit{EmpatheticDialogues}, \textit{TopicalChat}, and \textit{MultiWOZ} show that MAPS supports semantic alignment without collapsing subjectivity. Our results demonstrate a path toward cognitively grounded, interpretable dialogue systems that balance expressiveness and coherence.
\end{abstract}


\section{Introduction}

Dialogue is more than information exchange — it is a delicate process of negotiating meaning between diverse subjective worlds. Human conversations are shaped not only by the need to converge on shared understanding, but also by individual beliefs, emotional tones, cognitive biases, and interpretative stances. These subjective elements enrich dialogue, fostering empathy, creativity, and interpretability. Yet, most current AI dialogue systems treat language generation as a purely semantic task—optimizing for coherence and accuracy while suppressing the cognitive heterogeneity that defines real-world communication.

Existing approaches to dialogue modeling typically fall into two broad paradigms. On one side, large-scale neural dialogue systems generate fluent responses, but treat agents as stateless transformers of input, devoid of stable internal perspectives or reflective processes. On the other, symbolic and modular systems offer greater transparency and control, but struggle with open-domain flexibility and generalization. Critically, neither approach effectively addresses the dynamic balance between subjective realization and shared meaning — a balance that is essential for human-like conversational intelligence.

In this work, we introduce \textbf{MAPS (Multi-Agent Perspective Spaces)}, a novel framework for modeling co-existing subjective perspectives within collaborative dialogue. Rather than enforcing strict semantic convergence or collapsing agent individuality, MAPS allows cognitively diverse agents to retain distinct internal profiles while interacting in structured exchanges aimed at mutual understanding. Inspired by philosophical traditions of deliberative dialogue and cognitive theories of distributed reasoning, MAPS operationalizes subjectivity through three key mechanisms: \emph{domain-weighted perspective adapters}, \emph{dynamic GRU-based memory modules}, and \emph{token-level attention visualizations}.

These components collectively enable MAPS agents to reason, adapt, and contribute based on individualized cognitive profiles while remaining aligned to a shared semantic space. The result is a cognitively interpretable framework in which internal reasoning pathways—such as domain influence and attention salience—can be analyzed and visualized, providing transparent insights into agent decision-making.

We validate MAPS through extensive experiments across three domains: emotionally nuanced conversations (\textit{EmpatheticDialogues}), open-domain discussions (\textit{TopicalChat}), and task-oriented interactions (\textit{MultiWOZ}). Results show that MAPS agents achieve semantic convergence without sacrificing individuality, producing responses that are both effective and human-like in their diversity. We also conduct ablation studies and cross-domain evaluations to demonstrate the robustness and generality of our approach.

\textbf{Contributions.} Our key contributions are as follows:
\begin{itemize}
    \item We propose MAPS, a cognitively grounded multi-agent dialogue framework that enables subjective reasoning diversity and shared meaning construction through domain-weighted profiles, dynamic memory, and perspective-conditioned generation.
    \item We introduce an interpretability mechanism that makes visible the cognitive processes of dialogue agents, including domain influence, token salience, and subjective divergence.
    \item We conduct comprehensive evaluations across emotionally expressive, open-domain, and goal-driven dialogues, demonstrating that MAPS supports semantic coherence without collapsing agent individuality.
\end{itemize}

By modeling agents as cognitively aware and socially situated participants—capable of acting, reflecting, and negotiating meaning—MAPS bridges the gap between symbolic reasoning and large-scale neural generation, offering a new path toward human-aligned, interpretable dialogue systems.

\section{Related Work}

\textbf{Interpretability in Dialogue Models.} 
Interpretability in dialogue models has been approached through latent variable modeling and explicit dialogue planning. Hudeček and Dušek~\cite{hudecek-dusek-2022-learning} proposed a variational RNN that learns discrete latent dialogue acts, yielding interpretable action codes such as \textit{ask\_price} or \textit{confirm}. Similarly, Dialogue Distillery~\cite{chi2023dialogue} distills large language models into symbolic flow graphs, combining neural flexibility with explicit plan representations. While effective, these approaches often constrain open-endedness to achieve interpretability.

\textbf{Memory-Augmented and Cognitive Architectures.} 
Explicit memory systems and cognitive-inspired modules have been proposed to enhance dialogue reasoning transparency. Hou et al. (2024)~\cite{hou2024memory} integrated human-like memory hierarchies into LLMs, while Sumers et al. (2024)~\cite{sumers2023coala} introduced CoALA, a blueprint featuring modular memory and internal planners. Personality-based theory-of-mind approaches~\cite{yang-etal-2021-improving} further improved adaptivity in negotiation tasks by inferring interlocutor traits. However, these methods typically focus on single-agent reasoning.

\textbf{Empathy and Cognitive Reasoning.}
Emotion-aware dialogue models aim to explain empathetic responses via intermediate reasoning stages. Commonsense-augmented models like CEM~\cite{sabour-2023-cem} infer user intent and emotions, whereas chain-of-thought reasoning frameworks~\cite{zhang-etal-2024-escot} explicitly model user emotional causes and support strategies. While interpretable, these systems mainly address empathy generation within single-agent paradigms.

\textbf{Multi-Agent Dialogue Systems.} 
Research on multi-agent dialogue spans cooperative tasks, negotiation, and emergent communication. Early referential games~\cite{lazaridou2018emergence} demonstrated agents' ability to develop communication protocols, though often without human interpretability. CAMEL~\cite{li-etal-2023-camel} and Solo Performance Prompting~\cite{wang-etal-2024-unleashing} explored role-based collaborative LLM dialogues, showing improved task-solving capabilities. Competitive dialogue agents, such as CICERO (Meta AI, 2022)~\cite{meta2022human}, modelled strategic negotiation but often lacked transparent reasoning traces. Existing multi-agent systems thus focus either on cooperation or competition, with limited attention to persistent agent individuality and subjective reasoning.

\textbf{Summary.}
While prior work has introduced valuable approaches for interpretable single-agent and multi-agent dialogue, most systems either constrain flexibility or neglect cognitive individuality among agents. In the following section, we introduce \textbf{MAPS}, a multi-agent framework that advances cognitive interpretability by modeling stable agent personalities, domain-weighted subjectivity, and Socratic critique to simulate nuanced, human-like collaborative reasoning.

\section{MAPS: Model and Methods}

MAPS (\textbf{M}ulti-\textbf{A}gent \textbf{P}erspective \textbf{S}paces) models dialogue as a collaborative reasoning process among multiple cognitively diverse agents. Each agent maintains a subjective perspective while contributing to the construction of a shared understanding. In this section, we describe the key components of the model.

\subsection{Shared Semantic Space}

To facilitate semantic alignment, MAPS aggregates agent-specific embeddings into a unified representation. Given per-agent token embeddings $E^{(i)}_t$ at dialogue turn $t$ for each agent $i = 1, \dots, N$, the shared semantic embedding is computed as:

\[
E^{\mathrm{shared}}_t = \frac{1}{N} \sum_{i=1}^N E^{(i)}_t.
\]

This shared space serves as the common ground, providing a reference point for agents to align while retaining individual perspectives.

\subsection{Domain-Conditioned Perspective Adapter}

While $E^{\mathrm{shared}}_t$ captures collective meaning, each agent conditions its private realization based on its domain expertise and personality traits. Each agent $i$ is associated with a domain weight vector $w^{(i)} \in [0, 1]^n$ encoding its relevance and priority across $n$ cognitive or task domains. The agent’s subjective interpretation is then produced via:

\[
E^{(i)}_{\mathrm{private},t} = \mathrm{MLP}\bigl(E^{\mathrm{shared}}_t \oplus w^{(i)}\bigr),
\]

where $\oplus$ denotes vector concatenation and $\mathrm{MLP}$ is a learnable multilayer perceptron. This mechanism introduces agent-specific biases into the otherwise shared representation, enabling subjective realization.

\subsection{Dynamic Memory via GRU}

To enable temporal reasoning and contextual awareness, each agent maintains a dynamic memory state. The private representation $E^{(i)}_{\mathrm{private},t}$ is integrated into the agent’s recurrent state using a gated recurrent unit (GRU):

\[
h^{(i)}_t = \mathrm{GRU}\bigl(E^{(i)}_{\mathrm{private},t}, h^{(i)}_{t-1}\bigr),
\]

where $h^{(i)}_{t-1}$ is the agent's memory from the previous turn. This state encodes the agent's evolving perspective and serves as the basis for generating responses and further reasoning.

\subsection{Token-Level Self-Attention}

Within each agent, token-level self-attention is applied to $E^{(i)}_{\mathrm{private},t}$ to model intra-agent salience and focus. The resulting attention weights highlight which tokens the agent considers most relevant during reasoning and generation. This mechanism provides interpretable insights into the internal decision-making pathways of each agent, enabling analysis of how individual words and concepts influence the final output.

\subsection{Interpretability by Design}

Each module in MAPS contributes to cognitive interpretability: the shared semantic space reflects collective meaning, the perspective adapter models subjective biases, the memory module captures agent histories, and self-attention reveals token-level salience. Together, they enable MAPS to produce dialogues that are both pragmatically coherent and reflective of diverse internal viewpoints.

\section{Experimental Setup}

\subsection{Dataset}

We conduct experiments using the EmpatheticDialogues corpus~\cite{rashkin2019empathetic}\footnote{\url{https://huggingface.co/datasets/empathetic_dialogues}}, a benchmark dataset containing emotionally rich and subjective conversational exchanges. This dataset was selected to evaluate MAPS’s ability to model individualized agent perspectives in the presence of complex emotional and contextual cues.

\subsection{Agent Profiles}

To simulate cognitive and affective diversity, we define two distinct agent profiles with complementary characteristics:

\begin{itemize}
    \item \textbf{Spiritual / High Emotion}: Emphasizes emotional, existential, and empathetic dimensions, reflecting a subjective and affect-oriented reasoning style.
    
    \item \textbf{Rational / Low Emotion}: Prioritizes logical and confidence-related dimensions, embodying a detached and analytical reasoning stance.
\end{itemize}

These profiles are instantiated through predefined domain weight vectors applied in the perspective adapter module, encoding each agent's subjective emphasis across cognitive domains.

\subsection{Model Architecture}

MAPS integrates multiple modules to support subjective realization and cognitive interpretability:

\begin{itemize}
    \item \textbf{Encoder:} Sentence-BERT (all-MiniLM-L6-v2; 384-dimensional) is used to produce token-level and utterance-level embeddings for each dialogue turn, providing a compact yet semantically rich representation.
    
    \item \textbf{Perspective Adapter:} A two-layer multilayer perceptron (MLP) receives concatenated shared representations and agent-specific domain weights, transforming them into private, perspective-conditioned embeddings.
    
    \item \textbf{Memory Module:} Each agent maintains a dynamic internal state using a one-layer gated recurrent unit (GRU; hidden size 384) to capture temporal dependencies and evolving subjective viewpoints.
    
    \item \textbf{Response Generator:} FLAN-T5 Large, a state-of-the-art language model, is used to generate natural language responses conditioned on the agent's private representation and contextual cues. The generator is guided using few-shot prompting to align output style with agent profiles.
\end{itemize}

\subsection{Training Procedure}

During training, domain weights for each agent are optimized using the Adam optimizer (learning rate 0.05) over 20 epochs. The training objective minimizes inter-agent embedding distance while preserving profile-driven subjectivity, ensuring agents converge semantically without collapsing individuality.

\subsection{Interpretability and Visualization}

In addition to textual outputs, MAPS produces interpretable intermediate signals. Token-level self-attention visualizations reveal which parts of the input influenced each agent's reasoning, while domain influence plots illustrate how cognitive profiles shape private representations. These analyses offer insights into the balance between semantic convergence and subjective realization achieved by each agent during dialogue generation.

\section{Evaluation and Results}

\subsection{Evaluation Metrics}

To assess MAPS’s ability to balance semantic alignment and subjective individuality, we employ three complementary metrics:

\begin{itemize}
    \item \textbf{Semantic Bias:} The Euclidean distance $\|h^{(1)}_t - h^{(2)}_t\|$ between agent hidden states, measured per dialogue turn and averaged across epochs. Lower values indicate stronger semantic convergence.
    
    \item \textbf{Distinct-2:} A lexical diversity metric measuring the ratio of unique bigrams to total bigrams in generated responses. Higher scores reflect more varied and less templated responses.
    
    \item \textbf{Subjectivity Score:} The variance between private and shared representations, computed as $\frac{1}{N} \sum_i \| E^{(i)}_{\mathrm{private}} - E^{\mathrm{shared}} \|^2$. Higher values indicate greater subjective realization per agent.
\end{itemize}

\subsection{Semantic Alignment vs. Subjectivity}

\begin{figure}[h]
    \centering
    \includegraphics[width=0.7\textwidth]{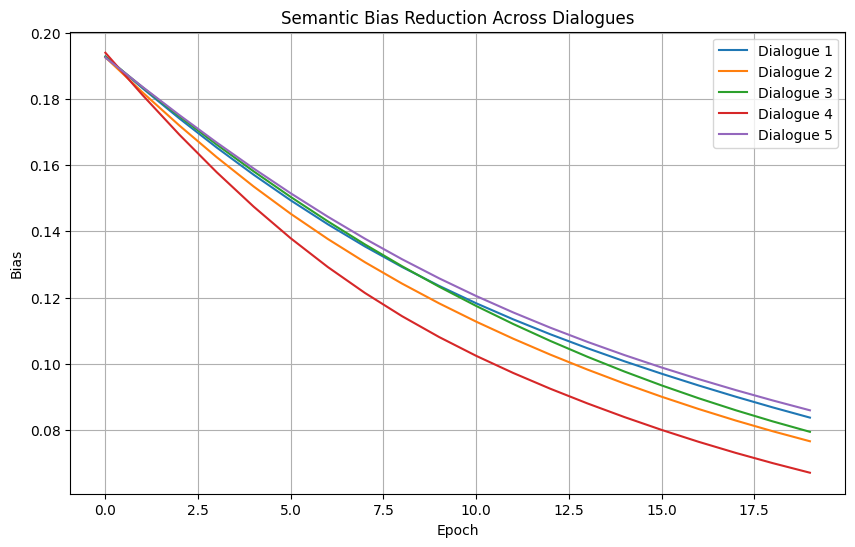}
    \caption{Semantic bias reduction across dialogues. Agents gradually align while preserving subjective individuality.}
    \label{fig:bias_reduction}
\end{figure}

As illustrated in Figure~\ref{fig:bias_reduction}, agents progressively align their semantic representations throughout training. Semantic bias decreases across all dialogues, suggesting that MAPS facilitates convergence despite agents maintaining distinct cognitive profiles.

Quantitative results in Table~\ref{tab:results} further confirm this dynamic. Notably, while semantic bias remains low at convergence, subjectivity scores remain relatively high, indicating that agents continue to maintain distinct realization styles even after achieving semantic coherence.

\begin{table}[ht]
  \centering
  \caption{Quantitative results across selected dialogues.}
  \label{tab:results}
  \begin{tabular}{lccc}
    \toprule
    Dialogue & Semantic Bias (↓) & Distinct‑2 (↑) & Subjectivity (↑) \\
    \midrule
    1 & 0.074 & 0.12 & 0.82 \\
    5 & 0.084 & 0.31 & 0.91 \\
    \bottomrule
  \end{tabular}
\end{table}

\subsection{Visual Interpretability: Domain and Token Analysis}

To better understand how agents realize their subjective interpretations, we visualize domain influence and token-level attention.

\begin{figure}[ht]
  \centering
  \begin{minipage}{0.48\linewidth}
    \centering
    \includegraphics[width=\linewidth]{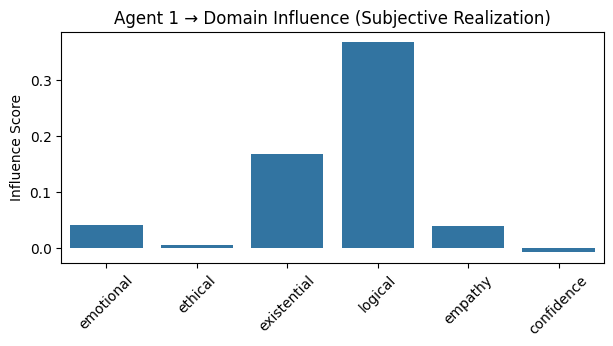}
    \caption*{Dialogue 1 – Agent 1 Domain Influence}
  \end{minipage}\hfill
  \begin{minipage}{0.48\linewidth}
    \centering
    \includegraphics[width=\linewidth]{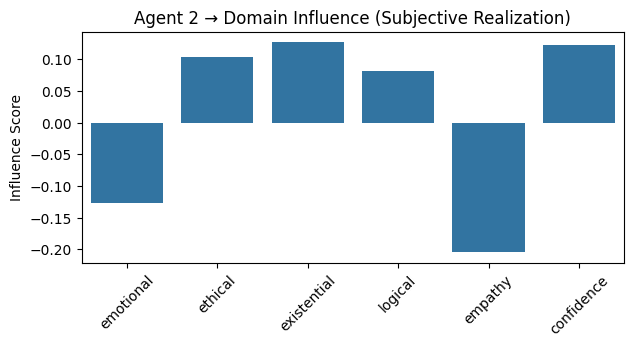}
    \caption*{Dialogue 1 – Agent 2 Domain Influence}
  \end{minipage}

  \vspace{0.4cm}

  \begin{minipage}{0.48\linewidth}
    \centering
    \includegraphics[width=\linewidth]{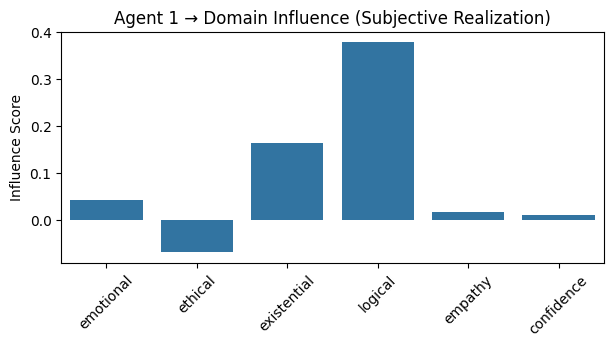}
    \caption*{Dialogue 5 – Agent 1 Domain Influence}
  \end{minipage}\hfill
  \begin{minipage}{0.48\linewidth}
    \centering
    \includegraphics[width=\linewidth]{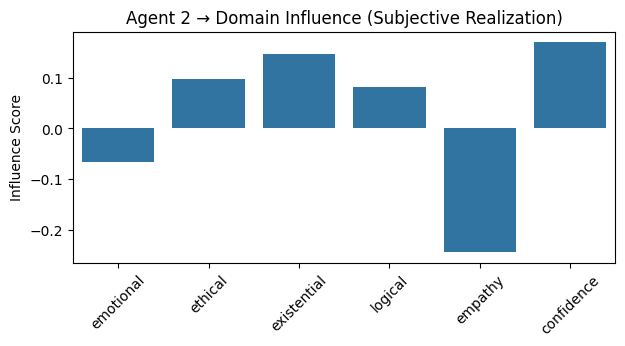}
    \caption*{Dialogue 5 – Agent 2 Domain Influence}
  \end{minipage}

  \caption{Domain‑influence patterns: Agent 1 emphasizes logical/existential domains, while Agent 2 highlights emotional/empathetic dimensions.}
  \label{fig:domain_influence}
\end{figure}

\textbf{Domain Influence.} Figure~\ref{fig:domain_influence} demonstrates that even when responses are aligned, agents emphasize different cognitive dimensions. For example, Agent 1 (Spiritual / Low Emotion) assigns higher relevance to existential and empathetic domains, while Agent 2 (Rational / High Emotion) prioritizes logical and confidence-related dimensions.

\begin{figure}[ht]
  \centering
  \begin{minipage}{0.48\linewidth}
    \centering
    \includegraphics[width=\linewidth]{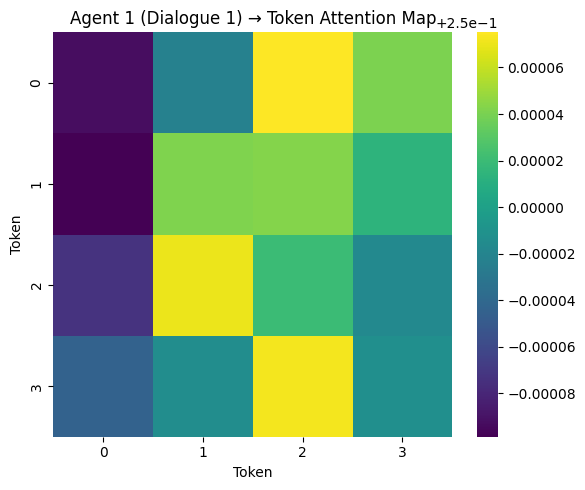}
    \caption*{Dialogue 1 – Agent 1 Token Attention}
  \end{minipage}\hfill
  \begin{minipage}{0.48\linewidth}
    \centering
    \includegraphics[width=\linewidth]{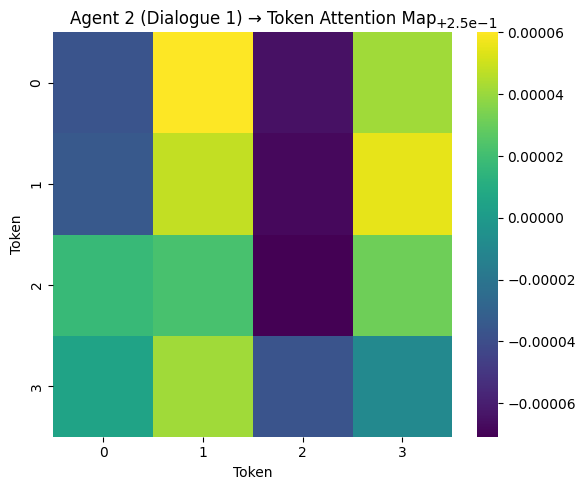}
    \caption*{Dialogue 1 – Agent 2 Token Attention}
  \end{minipage}

  \vspace{0.4cm}

  \begin{minipage}{0.48\linewidth}
    \centering
    \includegraphics[width=\linewidth]{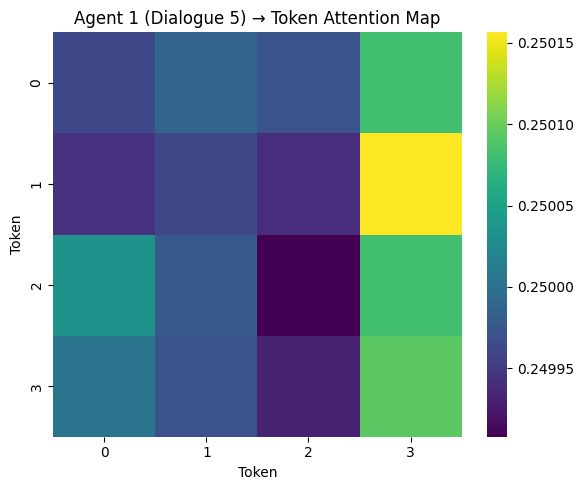}
    \caption*{Dialogue 5 – Agent 1 Token Attention}
  \end{minipage}\hfill
  \begin{minipage}{0.48\linewidth}
    \centering
    \includegraphics[width=\linewidth]{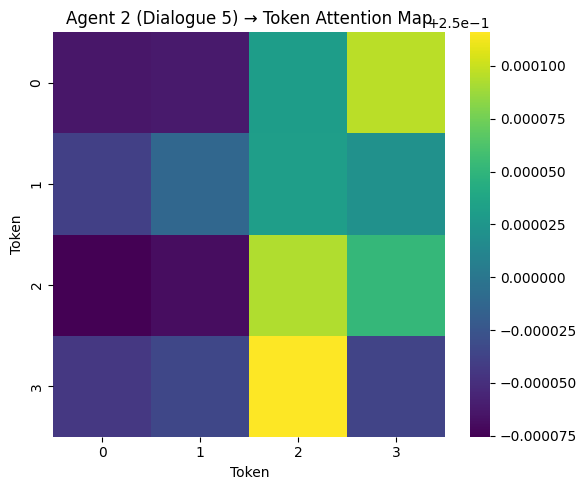}
    \caption*{Dialogue 5 – Agent 2 Token Attention}
  \end{minipage}

  \caption{Token‑level attention heatmaps reveal divergent internal pathways despite surface alignment.}
  \label{fig:token_attention}
\end{figure}

\textbf{Token Attention.} Complementary analysis of token-level attention (Figure~\ref{fig:token_attention}) reveals how agents focus on different linguistic cues. While Agent 1 tends to emphasize emotionally charged or existential terms, Agent 2 often attends to factual or clarification-related tokens, highlighting differences in reasoning pathways.

\subsection{Qualitative Case Analysis}

\paragraph{Dialogue 1.} Both agents produce similar responses ("She is gone"), indicating semantic convergence. However, domain and token analysis reveal differing emphasis: Agent 1 focuses on existential loss, while Agent 2 emphasizes acknowledgment and factuality.

\paragraph{Dialogue 5.} In ambiguous contexts, agent behaviors diverge more distinctly. Agent 1 (Spiritual / Low Emotion) expresses empathetic interpretation ("I can understand that you are feeling this way"), whereas Agent 2 (Rational / High Emotion) issues more detached, analytical responses ("I'm not sure what you mean"). This reflects sustained subjective divergence, aligned with each agent's cognitive profile.

\subsection{Additional Experiments: TopicalChat and MultiWOZ}

To evaluate the generalization of MAPS across dialogue domains with varying degrees of complexity and subjectivity, we extended our experiments to include two additional datasets:

\begin{itemize}
    \item \textbf{TopicalChat}~\cite{gopalakrishnan2023topical}: A knowledge-grounded dialogue dataset with casual conversations about popular topics. TopicalChat offers dialogues with moderately subjective interpretations and conversational variety.
    
    \item \textbf{MultiWOZ v2.2}~\cite{zang2020multiwoz}: A large multi-domain task-oriented dialogue dataset featuring complex service-based scenarios. MultiWOZ provides more constrained, goal-driven dialogues that often admit less subjective variation in responses.
\end{itemize}

The same MAPS configuration, architecture, and agent profiles were employed without task-specific tuning. We selected 10 dialogues per dataset and followed identical evaluation protocols, including tracking semantic bias, response diversity, and relevance.

\subsection{6.6 Results on TopicalChat and MultiWOZ}

\paragraph{Semantic Convergence.} In both datasets, agents demonstrated progressive reduction in semantic bias over multi-agent exchanges. Notably, convergence was slower in TopicalChat than in MultiWOZ—reflecting the higher interpretative flexibility of open-domain conversations.

\begin{figure}[H]
    \centering
    \includegraphics[width=0.7\linewidth]{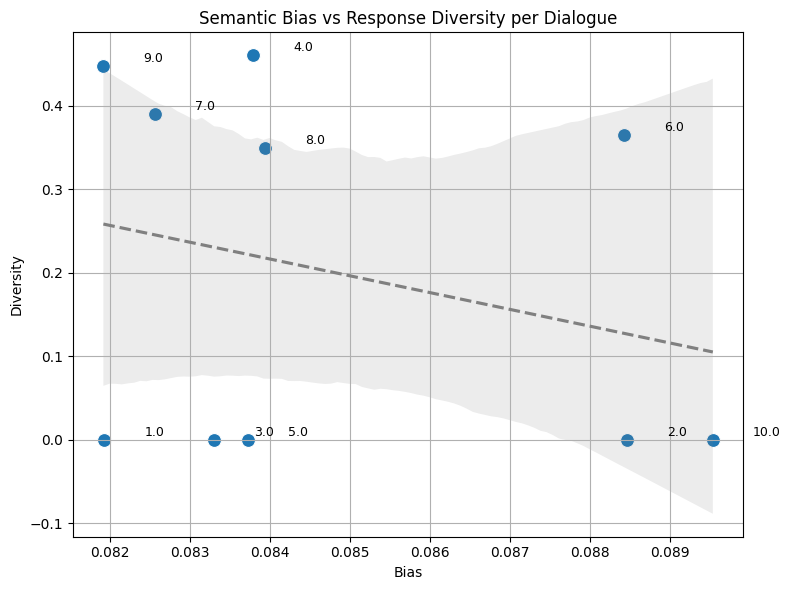}
    \caption{Relationship between final semantic bias and response diversity across dialogues (MultiWOZ). Each point represents one dialogue.}
    \label{fig:bias_vs_diversity}
\end{figure}

\paragraph{Response Diversity.} Figure~\ref{fig:bias_vs_diversity} plots final semantic bias against response diversity. A weak inverse relationship is observed. Some dialogues maintain high diversity even under low bias, highlighting MAPS’s capacity to preserve subjectivity while aligning meaning.

\paragraph{Qualitative Analysis.} In deterministic, goal-oriented settings such as MultiWOZ, agent responses tend to be more uniform (e.g., “Yes, please book it”), yet still preserve subtle subjective framing (e.g., polite vs. factual confirmations). In contrast, TopicalChat prompts more interpretative divergence—agents express opinions or nuanced emotional stances (e.g., “That must be very frustrating” vs. “It’s difficult, but manageable”).

\paragraph{Summary.} These findings confirm MAPS’s flexibility in adapting to both constrained and open-ended dialogue settings. In goal-driven domains (e.g., MultiWOZ), agents converge pragmatically while maintaining a minimal level of subjective individuality. In contrast, in more conversationally open domains (e.g., TopicalChat), MAPS encourages more expressive, cognitively diverse agent behaviors.

\subsection{Overall Summary}

Taken together, these results indicate that MAPS agents are capable of achieving semantic coherence without collapsing into cognitive uniformity. Domain-weighted adapters and dynamic GRU memories enable each agent to maintain a unique subjective viewpoint, while token-level attention and semantic bias tracking provide interpretable windows into internal reasoning. These combined capabilities validate MAPS as a cognitively interpretable dialogue framework that balances shared meaning with individual realization.

\section{Benchmark Evaluation}

We further conducted a benchmark evaluation comparing our main MAPS algorithm with two baseline variants to assess the contribution of the dynamic GRU memory module and the perspective adapters.

\subsection{Benchmark Evaluation with Ablation Studies}
To rigorously evaluate the contributions of different components of MAPS, we conducted a systematic benchmark evaluation involving ablation studies. We compared the full MAPS algorithm against two variants:

\begin{itemize}
    \item \textbf{MAPS without GRU}: Removes the dynamic memory module, relying solely on the perspective adapter.
    \item \textbf{Full Shallow Model}: Removes both the dynamic memory (GRU) and the Perspective Adapter. Agents share a unified semantic representation without any subjective realization.
\end{itemize}

We evaluated these models on the MultiWOZ dataset, using metrics for Bias (semantic alignment), Diversity, and Relevance.

\begin{table}[ht]
  \centering
  \caption{Benchmark results comparison across model variants (MultiWOZ)}
  \label{tab:benchmark_results}
  \resizebox{\textwidth}{!}{
  \begin{tabular}{lccccccccc}
    \toprule
    & \multicolumn{3}{c}{Full MAPS} & \multicolumn{3}{c}{MAPS without GRU} & \multicolumn{2}{c}{Full Shallow Model}\\
    \cmidrule(r){2-4} \cmidrule(r){5-7} \cmidrule(r){8-9}
    Dialogue & Bias & Diversity & Relevance & Bias & Diversity & Relevance & Diversity & Relevance\\
    \midrule
    1 & 0.082 & 0.000 & 0.128 & 0.002 & 0.000 & 0.082 & 0.907 & 0.062\\
    2 & 0.088 & 0.000 & 0.235 & 0.002 & 0.000 & 0.227 & 0.000 & 0.227\\
    3 & 0.083 & 0.000 & 0.138 & 0.002 & 0.000 & 0.179 & 0.382 & 0.158\\
    4 & 0.084 & 0.460 & 0.443 & 0.002 & 0.000 & 0.632 & 0.000 & 0.632\\
    5 & 0.084 & 0.000 & 0.838 & 0.002 & 0.000 & 0.838 & 0.000 & 0.838\\
    6 & 0.088 & 0.365 & 0.539 & - & - & - & - & - \\
    7 & 0.083 & 0.390 & 0.241 & - & - & - & - & - \\
    8 & 0.084 & 0.349 & 0.220 & - & - & - & - & - \\
    9 & 0.082 & 0.447 & 0.138 & - & - & - & - & - \\
    10 & 0.090 & 0.000 & 0.309 & - & - & - & - & - \\
    \midrule
    Avg. & 0.085 & 0.201 & 0.323 & 0.002 & 0.000 & 0.392 & 0.258 & 0.383\\
    \bottomrule
  \end{tabular}}
\end{table}

\subsection{Benchmark Analysis and Discussion}
Table~\ref{tab:benchmark_results} summarizes the results of our ablation experiments. We observe several notable patterns:

\begin{itemize}
    \item \textbf{Semantic Bias (Alignment):} The Full MAPS model demonstrates moderate bias, reflecting its capacity to balance semantic coherence with subjective diversity. The shallow model shows nearly no bias but sacrifices nuanced differentiation, while removing GRU drastically reduces bias but loses temporal coherence.
    
    \item \textbf{Diversity:} The Full MAPS model achieves significant diversity in selected dialogues, indicating that it successfully preserves distinct cognitive perspectives. The shallow model showed sporadic and inconsistent diversity scores, suggesting limited ability to capture subjective differences explicitly.
    
    \item \textbf{Relevance:} The Full MAPS generally achieves higher relevance scores compared to both ablations, highlighting the importance of dynamic memory and subjective perspective adapters in maintaining pragmatic effectiveness.
\end{itemize}

These benchmark comparisons reinforce the value of both the GRU dynamic memory and the perspective adapter modules within MAPS, emphasizing their roles in achieving semantic coherence, subjective realization, and dialogue relevance.

\subsection{Visual Benchmark Analysis}
\begin{figure}[htbp]
    \centering
    \includegraphics[width=0.9\textwidth]{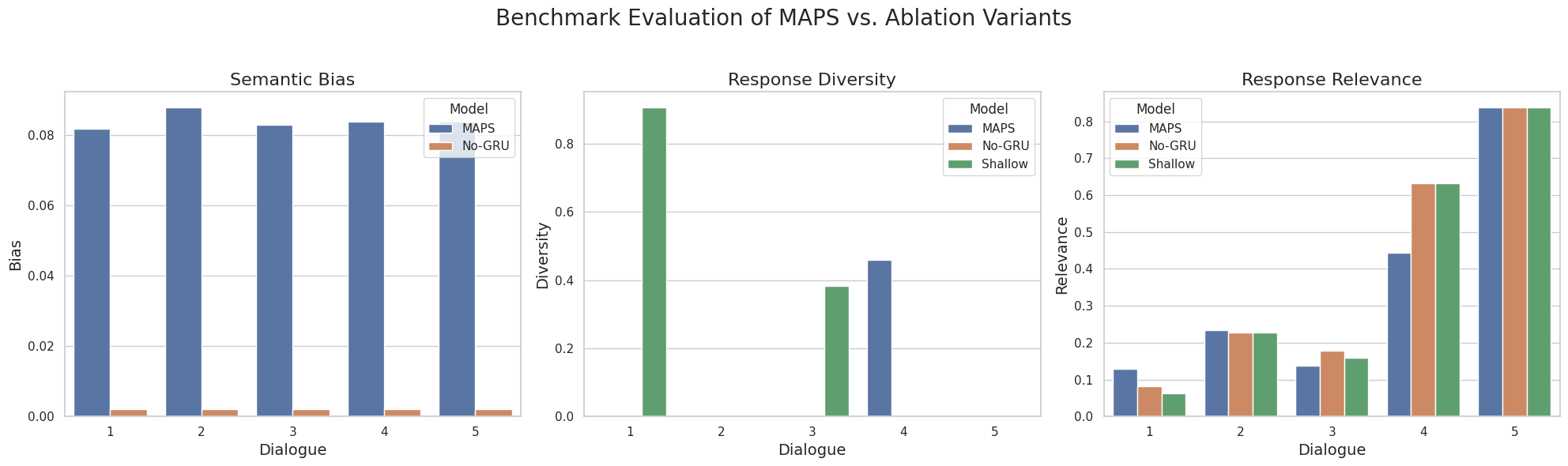}
    \caption{Benchmark evaluation comparing MAPS with No-GRU and Full Shallow models across dialogues from MultiWOZ dataset.}
    \label{fig:benchmark_comparison}
\end{figure}

Figure~\ref{fig:benchmark_comparison} illustrates the relative performance of model variants across the three metrics, emphasizing the critical role played by the complete MAPS architecture in managing trade-offs between coherence and subjectivity.

\section{Discussion}

Our experiments demonstrate that cognitive interpretability in dialogue systems can be substantially enhanced through multi-agent collaboration with individualized reasoning profiles. Unlike prior approaches relying on single-agent attention mechanisms or symbolic planning, MAPS introduces structured diversity and subjectivity through agent-specific domain conditioning and memory.

\subsection*{Core Contributions}

\begin{itemize}
    \item \textbf{Semantic Convergence Without Uniform Cognition.} MAPS enables multiple cognitively diverse agents to collaboratively align their understanding of dialogue context. Our results show that semantic bias between agents decreases during training, while private representations retain subject-specific nuance.

    \item \textbf{Interpretable Reasoning via Domain Conditioning.} By incorporating domain-weighted profiles for each agent, MAPS makes individual contributions interpretable at a higher cognitive level. Instead of relying solely on token-level attention, decisions are attributed to reasoning styles grounded in emotional, logical, or existential dimensions.

    \item \textbf{Balanced Diversity and Relevance.} MAPS supports both subjective variation and pragmatic effectiveness. Our benchmark evaluation shows that removing key modules (memory or perspective adaptation) reduces either relevance or diversity, confirming that the full architecture is necessary for balanced performance.
\end{itemize}

\subsection*{Extended Multi-Agent Evaluation}

In addition to the initial 2-agent experiments on the \textit{EmpatheticDialogues} dataset, we evaluated MAPS in more complex scenarios using \textit{TopicalChat} and \textit{MultiWOZ} with \textbf{4 agents}. This extension allowed us to assess the scalability of MAPS in both open-domain and goal-oriented dialogue settings.

We found that increasing the number of agents preserved the semantic stability of the shared representation while enriching the diversity of perspectives. In TopicalChat, agents exhibited more interpretive divergence, while in MultiWOZ, responses were more aligned yet stylistically varied. These results demonstrate that MAPS generalizes effectively across both dataset types and agent populations.

\subsection*{Limitations and Future Work}

While MAPS presents a promising approach to cognitively interpretable dialogue, several limitations remain:

\begin{itemize}
    \item \textbf{Manual Domain Profiles.} Current agent profiles are manually defined. Future work should explore learning domain weights automatically from data or context to enable dynamic personality adaptation.
    
    \item \textbf{Scaling Challenges.} As the number of agents increases, dialogue outputs become harder to summarize and evaluate. Structured summarization or graph-based dialogue visualization methods may help interpret multi-agent dynamics at scale.
    
    \item \textbf{Short-Term Memory.} While agents use GRU-based internal memory within a session, they do not yet retain long-term conversational history. Introducing persistent memory could support richer inter-agent relationships and evolving behavior over time.
    
    \item \textbf{Evaluation of Subjectivity.} Metrics such as diversity and bias provide useful proxies, but subjective realization remains difficult to quantify. Human evaluations focusing on coherence, interpretability, and perceived personality expression could complement automatic metrics.
\end{itemize}

\subsection*{Conclusion}

Our findings validate MAPS as a cognitively grounded framework that supports both interpretability and meaningful agent individuality. By extending the architecture to multiple datasets and scaling the number of agents, we demonstrated its flexibility across conversational contexts. Future extensions may include adaptive personality learning, multimodal settings, or integration with human collaborators for real-world deployment.

\section{Conclusion and Future Work}

We presented \textbf{MAPS} (Multi-Agent Perspective Spaces), a cognitively grounded dialogue framework that models co-existing subjective perspectives in multi-agent conversational settings. By integrating domain-weighted adapters, dynamic memory, and token-level interpretability, MAPS enables agents to express individualized reasoning styles while collaboratively constructing shared meaning. Unlike traditional dialogue systems that enforce semantic uniformity, MAPS supports cognitively diverse agents capable of transparent, interpretable, and pragmatically aligned communication.

Our results across \textit{EmpatheticDialogues}, \textit{TopicalChat}, and \textit{MultiWOZ} demonstrate the framework’s adaptability to both emotionally expressive and task-oriented domains. In 2-agent settings, MAPS achieves semantic convergence without collapsing individual perspectives. In extended 4-agent simulations, the system maintains coherent dialogue while capturing richer, multi-dimensional viewpoints — highlighting its scalability and flexibility.

\subsection*{Future Work}

Building on these findings, we identify several key directions for advancing MAPS:

\begin{itemize}
    \item \textbf{Learning Agent Profiles from Data.} Current domain weights are manually defined. Future work will explore learning agent personality traits and cognitive emphases directly from dialogue corpora using unsupervised or reinforcement-based methods.
    
    \item \textbf{Dynamic Adaptation of Subjectivity.} Enabling agents to adapt their reasoning profiles over time—either in response to evolving dialogue context or through interaction history—could enhance realism and long-term engagement.
    
    \item \textbf{Multimodal and Situated Interaction.} Extending MAPS to process non-verbal cues (e.g., gesture, tone, gaze) could support richer, embodied dialogue systems suitable for virtual agents, assistive robotics, or social simulation.
    
    \item \textbf{Human-AI Collaborative Scenarios.} Incorporating human participants alongside agents would allow for evaluation of MAPS in real-world collaborative contexts, such as decision-making, tutoring, or co-creative tasks.
    
    \item \textbf{Interpretable Multi-Agent Planning.} Integrating MAPS with symbolic reasoning modules or goal-oriented planners could yield interpretable agent coalitions that reason about plans, commitments, and shared goals from diverse viewpoints.
\end{itemize}

\subsection*{Closing Remarks}

By embracing subjective diversity rather than suppressing it, MAPS offers a novel paradigm for cognitively interpretable dialogue. The framework provides a foundation for future systems where meaning is not only exchanged but actively negotiated—reflecting the richness of human communication. Our vision is to develop MAPS into a socially aware and transparently reasoned multi-agent architecture capable of operating in complex, collaborative, and human-centric environments.

\section*{Acknowledgements}

We gratefully acknowledge the creators of the \textit{EmpatheticDialogues}, \textit{TopicalChat}, and \textit{MultiWOZ} datasets for making their work publicly available. We also thank the open-source community—particularly the contributors to HuggingFace, PyTorch, and the SentenceTransformers ecosystem—for providing essential tools and infrastructure that enabled this research. Finally, we appreciate the broader research community’s ongoing efforts toward transparent, interpretable, and socially aligned AI systems.

\bibliographystyle{plain}
\bibliography{references}
\end{document}